# An Ensemble-based approach for assigning text to correct Harmonized system code

Shubham
IBM Consulting

Avinash Arya.
IBM Consulting

Dr. Subarna Roy
IBM Consulting

Sridhar Jonnala
IBM Consulting

*Abstract*— **Industries must follow government rules and regulations around the world to classify products when assessing duties and taxes for international shipment. Harmonized System (HS) is the most standardized numerical method of classifying traded products among industry classification systems. A hierarchical ensemble model comprising of Bert-transformer, NER, distance-based approaches, and knowledge-graphs have been developed to address scalability, coverage, ability to capture nuances, automation and auditing requirements when classifying unknown text-descriptions as per HS method.**

## I. Introduction

Industries must follow government rules and regulations around the world to classify products when assessing duties and taxes for international shipment and for gathering statistics .Harmonized System (HS) is the most standardized numerical method of classifying traded products among industry classification systems.The HS is comprised of a hierarchically structured nomenclature which enables industries in US and its trading partner countries to classify their products till the 6-digit level following[1] published by world customs organization (WCO). Beyond this, countries are allowed to assign their own custom HS codes at 8–10-digit level. Taking an example from [2], the hierarchical structure of the code is as follows:

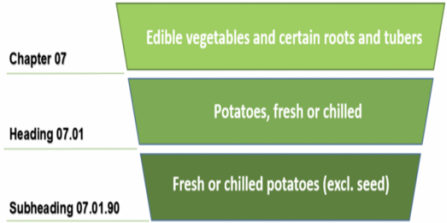

However, based on information from various third-party vendors that collects such data, it has been found that only 29% of the text that are entered by the users for shipments and computation of tariffs can be automated and assigned to the right HS code albeit with certain accuracy. This is because firstly, users are allowed to enter free text which generates messy /noisy/ non-sensical data at the backend. Secondly, to avoid higher tariffs or misuse exemption policies users often resort to entering ambiguous texts, making code assignment challenging for the system. Noisy and ambiguous text require manual intervention which is often error-prone leading to high degree of misclassification and revenue leakage Over the counter deep-learning text classification models or machine-translation models may not be of help in automating such texts. Based on data available in the public domain we have broadly classified such texts as follows:

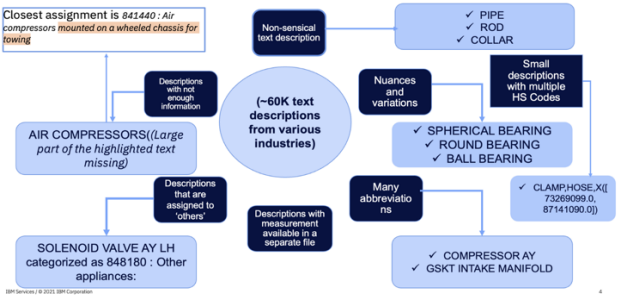

Figure-2 :*Types of text in a typical HS code classification system*

Also, a larger view of the document-hierarchy demonstrates the complexity of the classification problem. As one goes to the bottom of the hierarchy the degree of cardinality increases- rendering flat multi-label text-classification models unstable. They lack the capability to generalize as well.

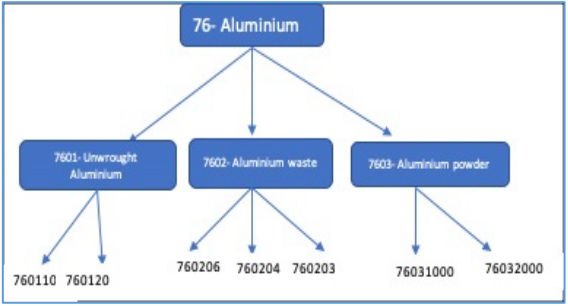

Figure-3 : *Hierarchical structure of the HS codes*

See, the above example, the level-1, describes the broad category (Aluminum and its products) which is at 2-digit level. The second level points to the set of categories of aluminum under 4-digit classification. The third level represents even granular level sub-headings (6-digit / 8-digit) of the respective categories. Often as we go to the bottom of the pyramid the volume distribution under the categories gets skewed with some of them being significantly large and others being thinly populated.

While, multi-label classification models are useful at 1st two levels, at more granular levels the following challenges are observed:

A classification model at XX-digit level (which provides unconditional probabilities for a text description to belong to a particular class at XX-digit) might require thinly populated classes to be grouped into 'other' category. Thus, new text description that has a potential to fall into this category will miss out on being labelled into a specific class- leading to manual intervention. If they are not grouped, then the model will lack generalization. For example, let's say 76.01.20 is a thin-volume class and hence has been grouped with rest of the thin volume classes forming a 'other' category during training. If a new text-description has a potential to belong to this class e.g., 76.01.20, no way from the model it can be determined that it would be the case.

Secondly, a flat classification model would miss out on exploring the possibilities of belonging to a specific class or set of classes at each level. Flat model lacks the capability to traverse the search-space of each level and use that information in classification at the final level- thus eroding models' ability to capture nuances. Hence, model training should be hierarchical in nature. The theoretical framework for this kind of model training will be explained in section-III.

Also, since problem related to thin volume is more severe at lower end of the classification pyramid, supervised learning by leveraging companies own data should be restricted till the 4-digit level even when the hierarchical option of model training is exercised. At 6-digit level and beyond using a combination of unsupervised learning methods e.g., sentence-embedding, named entity recognition [NER], semantic search, cosine-similarity and knowledge graph can yield better result. Therefore, coverage of the system to classify more text-description in specific code increases.

We demonstrate the superiority of the later approach using publicly available dataset [3] based on several metrics – e.g., scalability, coverage, generalization, ability to capture nuances and variations in text etc. We further show how knowledge-graph can be designed based on the WCO manual. The proposed methodology is a useful utility to highlight missing information in user-provided text. This serves the audit requirement of providing rationale for HS classification.

## II. METHODOLOGY

*A. The Flat Model*

The outcome of training a flat model is the unconditional probability for a given feature vector $x_i$ extracted from i$^{th}$ description to belong to any one of the $K$ classes. Each class comprises of one 6-digit HS code. Let's denote the predicted class by $\hat{y}_i$, then the unconditional probability of it being equal to the j$^{th}$ class is given by the SoftMax function:

$$p(\hat{y}_i = j) = \frac{e^{x_{ij}}}{\sum_{k=1}^{K} e^{x_{ik}}}, \forall j = 1 \dots K \quad (1)$$

In this approach a pre-trained Bert – based sequence classifier is used. The output layer of the Bert model is replaced by the soft-max layer as per equation-1 to meet the requirement of multi-label classification. A typical architecture is represented below:

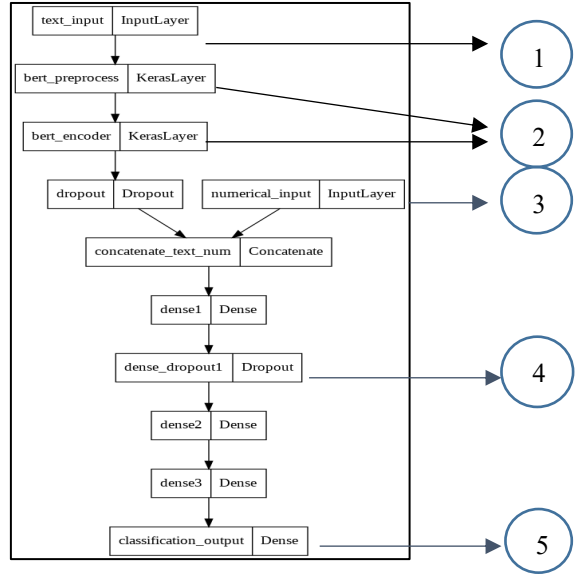

Figure-4: *(1)represents the input text layers. The input texts arrive in sequence and get pre-processed in the BERT layer represented by (2). The second Bert-layer in (2) creates the contextual embeddings which passes through the subsequent layers as feature vectors along with other numerical inputs as shown by (3). There is a drop-out layer at (4) to prevent over-fitting, The output-layer is represented by the soft-max function as represented by equation(1) above. The model is trained with the objective of minimizing the cross-entropy loss function.*

*B. The Hierarchical Model*

This modelling approach is a 3-step process. First, a 2-digit classifier is built in the same way as explained in subsection A for a 6-digit classifier. Let us say there are $K^{(2)}$ classes at 2-digit level. Hence, the unconditional probability of predicted 2-digit class $\hat{y}_i^{(2)}$ being equal to $l$ is given by:

$$p(\hat{y}_i^{(2)} = l) = \frac{e^{x_{il}}}{\sum_{k^{(2)}=1}^{K^{(2)}} e^{x_{ik^{(2)}}}}, \forall l = 1 \dots K^{(2)} \quad (2)$$

The same architecture as explained in Fig.4 is followed to obtain a 2-digit classifier for a set of text-descriptions. The 2-digit class with the highest probability of occurrence for the ith observation is obtained based on minimizing the cross-entropy loss function in (3)

$$C = \sum_i^n \sum_{k^{(2)}=1}^{K^{(2)}} -\hat{y}_i^{(2)} \log p(\hat{y}_i^{(2)} = l) \qquad (3)$$

Let, $l^*$ be the class with the highest probability of occurrence for ith observation

$$p(\hat{y}_i^{(4)} = l^*) = \max \cup_{k^{(2)}}^{K} p(\hat{y}_i^{(2)} = l)$$

In the next step, the probability of predicted 4-digit class $\hat{y}_i^{(4)}$ being equal to $m$ can be computed as,

$$p(\hat{y}_i^{(4)} = m) = \frac{e^{x_{im}}}{\sum_{k^{(4)}=1}^{K^{l^*(4)}} e^{x_{ik^{(4)}}}}, \forall m = 1 \ldots K^{l^*(4)} \qquad (4)$$

This is under the assumption that $K^{l^*(4)}$ is the number of 4-digit classes under $l^*th$ 2-digit class. Hence, the conditional probability that predicted 4-digit class $\hat{y}_i^{(4)}$ being equal to $m$ given predicted 2-digit class $\hat{y}_i^{(2)}$ is equal to $l^*$ can be written as:

$$p(\hat{y}_i^{(4)} = m | \hat{y}_i^{(2)} = l^*) = \frac{e^{x_{im}}}{p(\hat{y}_i^{(2)}=l)(\sum_{k^{(4)}=1}^{K^{(4)}} e^{x_{ik^{(4)}}})} \qquad (5)$$

Please note that in our current empirical exercise in section -III, the unconditional soft-layer at 4-digit level is applied based on (4). Also, for current implementation the training dataset for this level contains all observations that have historically been assigned a class $l^*$ and not only those for which $l^*$ is the class with highest probability of occurrence.

The 4-digit classifier is computed using (4), corresponding cross entropy function and the architecture in Fig.4.

At the third step, distance-based unsupervised modelling approaches are considered by comparing the organization's dataset at 6-digit level first with training set and then with the WCO manual. If the i[th] text-description falls in 4-digit class $m^*$ based on maximum probability of occurrence, then it is compared will all the 6-digit classes under $m^*$ in the training dataset using cosine similarity. Bert-based sentence embedder is leveraged to convert all the text-descriptions into vectors including the i[th] text-description description for which comparison is made. The cosine similarity between i[th] text-description and o[th] text-description from the training set is computed based on the following equation:

$$\rho(i, o) = \sum_j cosine(i, o) \qquad (6)$$

Let's say, $o^*$ is the class for which $\rho(i, o)$ is maximum. So $o^*$ is chosen as the 6-digit class for i[th] text-description.

The above approaches all rely on the training dataset to male predictions. But as demonstrated in figure-2 the historical dataset is highly susceptible to human-errors. Users have to fill-up a form while sending their shipment. The data created is the digital copy of what users enter manually in the form. Given the extent of data created every day, practically it is not possible to validate each row. So, over-reliance on training dataset to predict HS code in future would aggravate the problem of incorrect assignment rather than solving it.

Secondly, there is frequent changes in the WCO manual which makes the rules that were used for previous assignments invalid. Hence, over-reliance on training data under this changing circumstance will render HS code prediction almost ineffective. Therefore, at the third step of the methodology we also suggest comparison of the text with WCO manual to get alternative suggestions for HS code which might be more appropriate than what is assigned based on past examples.

The end-to-end methodology of comparison with WCO is described in the next subsections. In this case, all the 6-digit classes under $m^*$ in WCO is modelled as a knowledge-graph through custom rule extraction. Let's say, $m^*$ corresponds to the 4-digit class 84.14, Fig.5 demonstrates some of the 6-digits and 8-digits codes and standardized description under this class. The process-flow is as follows:

- Extraction of customs-rules from WCO descriptions
- Visualization the rules in the form of a knowledge-graph
- Comparison of entities in the 6-digit descriptions with nodes and links in the knowledge graph and compute average cosine -similarity
- Select the 6-digit class that corresponds to the maximum average cosine-similarity

| 8414 | | Air or vacuum pumps, air or other gas compressors and fans; ventilating or recycling hoods incorporating a fan, whether or not fitted with filters; parts thereof: |
|---|---|---|
| 8414.10.00 | 00 | Vacuum pumps . . . . . . . . . . . . . . . . . . . . . . . . . . . |
| 8414.20.00 | 00 | Hand- or foot-operated air pumps . . . . . . . . . . . . . . . |
| 8414.30 | | Compressors of a kind used in refrigerating equipment (including air conditioning): |
| 8414.30.40 | 00 | Not exceeding 1/4 horsepower . . . . . . . . . . . . . . . |
| 8414.30.80 | | Other . . . . . . . . . . . . . . . . . . . . . . . . . . . . . . . . . . |
| | | Screw type: |
| | 10 | Not exceeding 200 horsepower . . . . . . . . . |
| | 20 | Exceeding 200 horsepower . . . . . . . . . . . . |
| | | Other: |
| | | For all refrigerants except ammonia: |
| | 30 | For motor vehicles . . . . . . . . . . . . . . . |
| | | Other: |
| | 50 | Exceeding 1/4 horsepower but not exceeding 1 horsepower . . . . . |
| | 60 | Exceeding 1 horsepower but not exceeding 3 horsepower . . . . . |
| | 70 | Exceeding 3 horsepower but not exceeding 10 horsepower . . . . |
| | 80 | Exceeding 10 horsepower . . . . . . . |
| | 90 | For ammonia . . . . . . . . . . . . . . . . . . . . |
| 8414.40.00 | 00 | Air compressors mounted on a wheeled chassis for towing . . . . . . . . . . . . . . . . . . . . . . . . . . . |

Figure-5: *Descriptions for headings and sub-headings of 8414. The digits highlighted in red are either 6-digit classes or classes which have no further sub-headings post 6-digit*

## C. Custom-rule extraction

To extract entities and the relationships between them, we created a rule-based entity-relationship extraction algorithm which given a sentence splits it into entities and relationships. The sentence is first POS tagged and passed to the rule-based extractor which works upon the POS tags.

Through POS tagging, a dependency graph is generated which can be traversed using certain rules to identify entities and relationships.

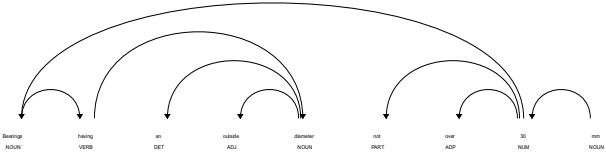

Figure-6: *Dependency graph of a POS tagged sentence*

## D. Knowledge-graph

Post segmentation of the text corresponding to a given 6-digit into meaningful phrases, the entities containing nouns/pronouns are represented as nodes. The contextual word to the nouns also goes to the relevant nodes. Rest of the entities are considered as linking phrases and assigned to the edges.

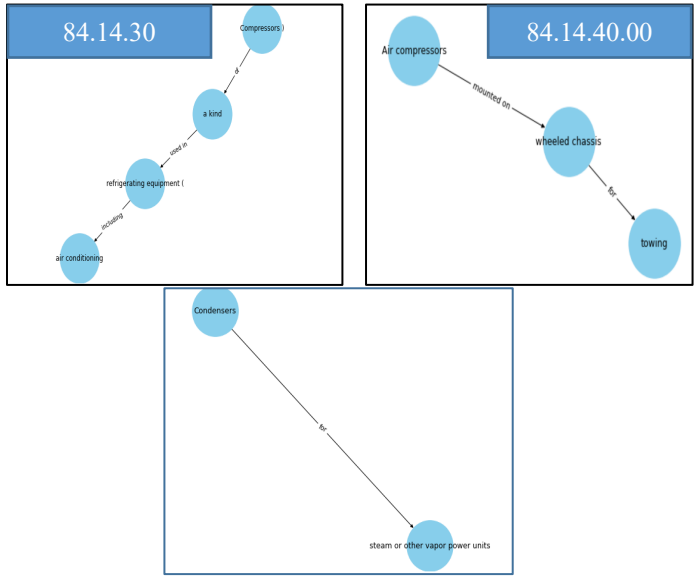

Figure-6: *Three Knowledge graph representations of descriptions for 84.14.40.00 and 84.14.30*

## E. Entity comparison with distance measures

The $i^{th}$ text that has a maximum probability of occurrence in the 4-digit class of $m^*$ is then compared with each node and links of the knowledge graph of all the associated 6-digit classes. Let's say there are $K^{m^*(6)}$ 6-digit classes under $m^*$th 4-digit class. If there are $S^o$ nodes and links within the $o$ th 6-digit class in accordance with the construction of the knowledge graph, then the average cosine similarity ($\rho^o$) is computed based on the following:

$$\rho^o(i,o) = \frac{\sum_j cosine(i, o^{(j)})}{S^o} \quad (6)$$

$o^{(j)}$ is the $j^{th}$ node/link of the $o^{th}$ 6-digit class that belongs to $m^*$th 4-digit class. Let's consider $o^*$ is the class for which $\rho^o$ is maximum. Then the $i^{th}$ observation will be assigned the $o^{*th}$ 6-digit class. For the task of execution of cosine similarity methods, first the sentences/phrases from product-description and HS code descriptions in WCO are vectorized using sentence-BERT base version.

In the implementation, we considered top 3 most probable HS codes in terms of average cosine similarity instead of only the top-most. The indicative high-level workflow is given below:

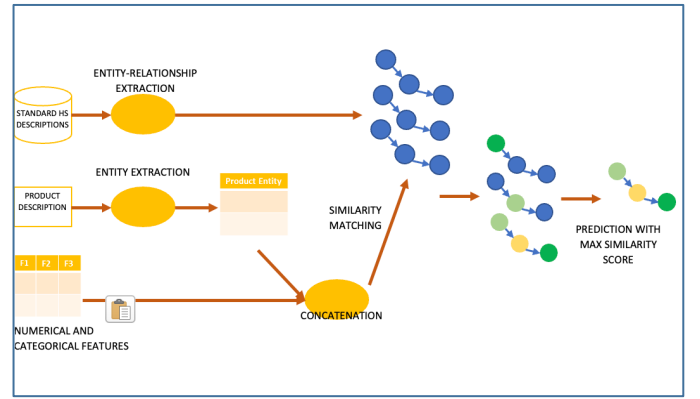

Figure-7: *Comparison of text based on cosine-similarity and knowledge-graph approach*

This approach is like [4], but one significant difference is the custom rule extraction layer that extracts entities and links from the HS code description. Each entity and link are then used to form the knowledge graph. The pre-processed product description is then compared with each of these entities/links to come up with the average cosine-similarity score. Fig.7 illustrates the approach where a pre-processed product description is compared with three knowledge-graphs, the nodes, and links of each of which is extracted from HS code description in WCO manual. One that has the highest match in terms of average cosine-similarity is chosen. Post comparison the color of each graph changes to green, light-green, and yellow depending on the extent of the match with the highest match represented by green. In case, there is no match the node or link color remains blue.

## III. EMPIRICAL APPROACH

### A. DATA

The dataset used for HS code classification is a publicly available data. This dataset contains cargo shipment description, harmonized code, and other features. Product descriptions having been assigned with 6+ digit HS code have been used for training Flat 6-digit model and Hierarchical model. Exogenous variables- harmonized weight and harmonized value have been used in model training. HS4 Classification model is trained on Chapter 84 since it is the class with the highest volume among all the 4-digit classes. TABLE 1. contains key details about train data and test data for flat and hierarchical model.

| Model | Train Data Size | Test Data Size | No. of classes | No of classes with volume greater than 5% |
|---|---|---|---|---|
| Flat (HS6) Model | 232467 | 77490 | 555 | 5 |
| Hierarchical (HS2) Model | 232467 | 77490 | 54 | 5 |
| Hierarchical (HS4) Model – Chapter 84 | 54471 | 18158 | 44 | 6 |

TABLE II. Model type and number of records used for training and testing the model for HS code classification.

As is evident from the table, very few classes have significant volume (> 5%), more so for HS6 classes, though the number of classes in this category is the highest, therefore many classes within it have negligible volume rendering direct classification ineffective.

*B. PRE-PROCESSING STEPS*

Before using the data for modelling, we apply some pre-processing steps which yields in better model performance:
The pre-processing steps vary according to the data. Some datasets have very concise and accurate descriptions, while some have descriptions which require a lot of cleaning and processing.

1. Text pre-processing steps:
    a. Remove extra-spaces, special characters which do not provide meaning to the description
    b. Remove Email IDs, phone numbers, fax numbers and other unnecessary texts
    c. As these are manually written descriptions, there might be incorrect spellings, we correct such words using various techniques to a certain extent of error.
    d. Descriptions also contain certain abbreviations which can be specific to that company alone. We replace certain known abbreviations with the actual word.
    e. Lemmatization

2. Ambiguous description data points:
    Certain descriptions in the dataset are assigned to multiple HS codes, which can lead to decreased model performance.
    Therefore, in a particular ambiguous description, if a HS code occurs more than 80% of the times, we retain that description-HS code combination and remove the rest.
3. Low frequency classes are grouped together to form a 'others' class.
4. Standardizing numerical features

IV. RESULTS

Post pre-processing, both flat and hierarchical models were trained as per the methodology described in section 2. The accuracy was computed with test sample of 300 observations. The results are provided below. There is a 16% increase in accuracy if hierarchical model is implemented vis-à-vis a flat model. The flat model also was unable to classify 12% of the text-descriptions into legitimate classes. While the proposed hierarchical model assigned valid classes to 100% of the text-descriptions.

Both the models were trained using transformers and demonstrated outstanding ability to capture nuances. For example, the flat model was able to rightly classify the text-description "casting parts finned aluminium air cleaner hb" in '76', though it was originally classified in '84'. Similarly, "mens polo style pm1001 style" was rightly classified as '62', though it was originally assigned to '84'. One more interesting result is the assignment of "pallet sodium trifluoroacetate 25 0kk net weight 100 0000 kgs ams hbl 1855728061 scac bopt" in '39' which is a correct assignment for chemicals, while it's original classification was '84'.

We also demonstrate the superior classification ability when texts are compared with knowledge-graph extracts of WCO manual for distance calculation. For example, "package stc conical roller bearings" has been originally assigned a HS code of 848291 which as per the WCO manual is referred to texts related to "*Balls, needles and rollers:*". However, when all related knowledge graphs are extracted based on methodology described in section II, the knowledge-graphs related to the HS codes (848250, 848251) appeared to be more appropriate.

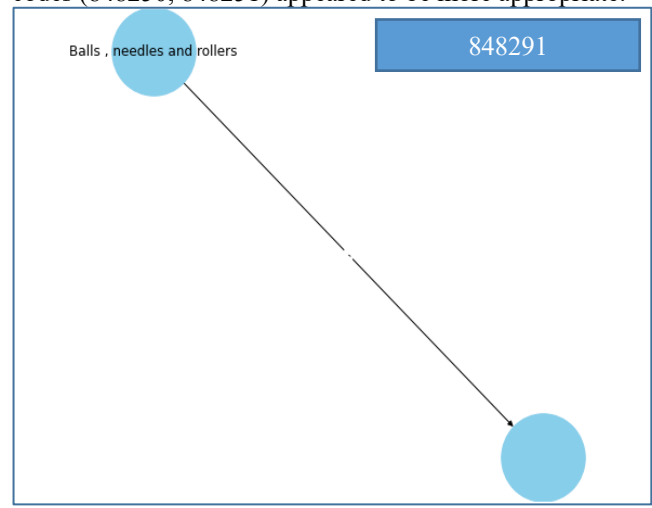

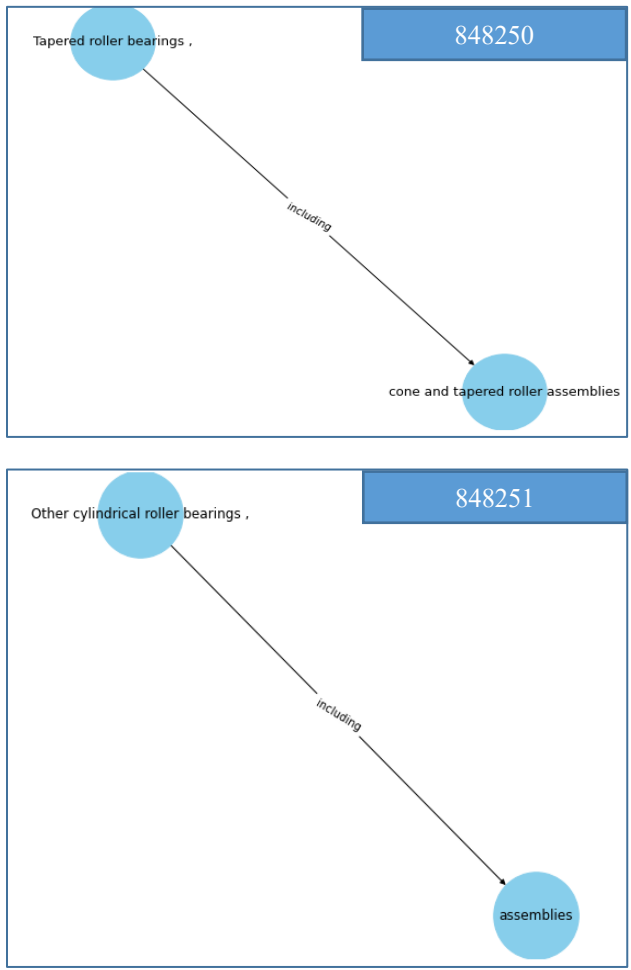

Figure 7: *Since, the node of 848250 has the keyword 'cone' in it, while 848251 has the keyword 'cylindrical' along with the keyword roller bearing, understandably, when the cosine similarity is computed these two codes got higher score than 848291. The code 848291 has only the keyword 'roller' in the node.*

We have discussed on approaches that would provide superior results when training dataset is not reliable or there is continuous update in HS classification codes at a global level and sometimes at individual country level.

## V. CONCLUSIONS

We have discussed and compared the two different approaches – the hierarchical and the flat approach to address the HS code classification problem. While the hierarchical methodology with its conditional probability approach ensures that texts get assigned to right code with a greater accuracy, imbalance in data translates greater biasedness towards low-level classification (HS2)- eventually impacting the hierarchical model as well. Such high imbalance is not visible for Flat Model. One way to address this problem is to build the overarching model of the hierarchical methodology at industry level rather than at a chapter level

Another important contribution of this paper is the interpretability through knowledge-graph. We have demonstrated the ability of the solution to show why a particular text is being assigned to a specific HS code- which would help in audit purpose. This approach also introduces flexibility in the solution and reduces sole dependency on training data.

## REFERENCES

[1] https://dl.acm.org/doi/fullHtml/10.1145/3468891.3468915,

[2] I. S. Jacobs and C. P. Bean, "Fine particles, thin films and exchange anisotropy," in Magnetism, vol. III, G. T. Rado and H. Suhl, Eds. New York: Academic, 1963, pp. 271–350.

[3] K. Elissa, "Title of paper if known," unpublished.

[4] R. Nicole, "Title of paper with only first word capitalized," J. Name Stand. Abbrev., in press.

[5] Y. Yorozu, M. Hirano, K. Oka, and Y. Tagawa, "Electron spectroscopy studies on magneto-optical media and plastic substrate interface," IEEE Transl. J. Magn. Japan, vol. 2, pp. 740–741, August 1987 [Digests 9th Annual Conf. Magnetics Japan, p. 301, 1982].

[6] M. Young, The Technical Writer's Handbook. Mill Valley, CA: University Science, 19